\documentclass[12pt]{article}
\date{}
\usepackage{epsfig}
\usepackage{amssymb,amsmath}
\usepackage[mathscr]{eucal}
\numberwithin{equation}{section} 

\usepackage{amssymb,amsmath, amsthm}
\usepackage[mathscr]{eucal}
\usepackage{amsfonts}
\usepackage{mathrsfs}
\usepackage{graphicx}

\newtheorem{proposition}{Proposition}

\title{Enhanced and Efficient Reasoning in Large Learning Models\\
}

\author{Leslie G. Valiant\\
John A. Paulson School of Engineering and Applied Sciences\\
Harvard University, Boston, MA 02134\\
{\it valiant@seas.harvard.edu}\\
}
\begin{document}
\date{May 12, 2026}
 \maketitle
\thispagestyle{empty}

\begin{abstract}

In current Large Language Models we can trust the production of smoothly flowing prose, which is justified in terms of the principles of machine learning. However, there is no comparably principled basis to justify trust in the content of the text produced. The widely recognized phenomenon of hallucinations is just one manifestation of this situation. It appears to be conventional wisdom that addressing this issue by adding more principled reasoning is not computationally affordable. 

Here we propose a principled method of reasoning that is efficient enough to be practical for large language models. Further, the method allows the retention of much of the currently used software and hardware base. Our method for improving the functioning of large language models consists of a first stage of preprocessing that recodes the data to be more explicit about the relationships among the objects described in the text, followed as a second stage by a standard but possibly streamlined machine learning process that then also learns to predict these relationships. This paper discusses a particular data recoding, the Unary Relational Integracode.

The method may be viewed as realizing a world model that applies beyond natural language, to vision and actions, for example, where the multiple properties of an object referred to in an input are brought together explicitly, rather than remaining distributed in the various references to it in the input.  We articulate its advantages in terms of Robust Logic, a system for performing principled chaining on learned, and hence uncertain, information.  We show that our recoding has the surprising and fortuitous property that, while succinct, it makes a core problem in learning relational rules that hold in the world described in the training data polynomial time learnable, the polynomial depending on the complexity of the rule. This gives heuristic support for sound reasoning {\it within} each call of the learned classifier. Additionally, the method gives similar support for reasoning {\it between} multiple calls, which are essential to large language models both for successive next token prediction and for more general reasoning methods. Each call of our method outputs explicit relational information. For multiple calls, therefore, the method offers some immediate efficiencies over current methods that lose between calls any relational information that they may have gleaned within calls.

\end{abstract}


\section{Heuristics in Computation}
The practice of computing relies on the use of heuristic algorithms to cope with situations where we believe there is no universal panacea. In the analysis of programs by compilers heuristics are needed since we know that many of the properties we may wish to detect, such as whether a program terminates, are not computable. In combinatorial optimization, for solving integer programming problems, for example, heuristics are used because it is the current belief that, since these problems are NP-complete, there is no efficient algorithm that will solve all cases. In machine learning similarly it is currently believed that not all data sources are PAC learnable, and over the last several decades a large number of learning algorithms have been intensively compared on large numbers of shared datasets to determine which ones are most effective, as heuristics, for generalization for various sources of data.

In all the above cases of program analysis, combinatorial optimization and machine learning, the use of heuristics is principled in the sense that in each case we can (i) define precisely what we {\it wish} to accomplish, and (ii) we have some rationale for believing that either in perpetuity or just for now we cannot have our wish realized in all cases.

By these standards, the development of large language models (LLMs) for producing good predictions of the next token can be considered to be principled as it follows the established paradigm of training and testing, whether on just text or on further examples of human preferences, in the sense of PAC learning  \cite{Valiant1984},\cite{KearnsValiant1994},\cite{Mohri2018}. Unfortunately, this is not the end of the story. First, we now ask more and more of LLMs. We expect their output to not only form readable well-formed paragraphs but also to {\it make sense} in the world. We expect LLMs to perform tasks for which they have not been trained as directly. Of course, for any very specific task and with enough effort, we can assemble a suitable training set. But this appears to work only one task at a time and any number of different tasks can be identified \cite{Hendrycks2025}. 

A more general view of the process of learning for the purpose of making sense of the world has been defined as {\it Knowledge Infusion} \cite{Valiant2006}. That definition requires, in particular, that the learned knowledge be ``in such a form that principled reasoning on it can be carried out computationally feasibly''.  Current LLMs have grown out of the chatbot paradigm \cite{Weizenbaum1966} but now there is a demand for them to fit the knowledge infusion paradigm. This paper observes that these two paradigms can be reconciled more efficiently than one might have thought, but some changes to current LLMs are needed. It is true that chatbots often produce outputs that make sense in the world and that are apparently consistent with correct reasoning, but their performance in this is widely recognized to lack reliability. We might aim for systems that are more dependable in functionality, in the sense spelled out in knowledge infusion. 

In this endeavor we also need to have an eye on energy consumption in addition to functionality. Currently used transformers \cite{Vaswani2017}  consume substantial energy. It is not known whether they are efficient for the functionality they achieve. We are suggesting a broader space of recodings and architectures to be explored {\it both for increased functionality and for efficient energy usage.} Given the scale of resources that humanity is planning to invest in AI, such investigations would appear to be warranted and timely. 

We suggest that to make LLMs more effective in reasoning, factuality, and absence of hallucinations, and at the same time as energy efficient as possible, we need a more refined understanding of the following Basic Question: {\it What provisions should we make to facilitate the processing of the interactions among different, possibly distant, parts of the text, balancing functionality with costs?} 

Currently deployed LLMs generally employ the transformer architecture with text input as the answer to this Basic Question. These use attention matrices to provide a generic mechanism for expressing levels of affinity between different pairs of tokens in a window of text. Here we suggest a different answer to the Basic Question. The underlying thesis is that the interaction information that needs to be extracted from text is of the same general nature as relationships among entities that have been considered as being needed to model cognitive processing in human working memory \cite{Quillian1968},\cite{NewellSimon1972}\cite{Anderson1973}. Robust Logic \cite{Valiant2000} is a quantitative computational model of this that emphasizes learnability. Central to it is a {\it Mind's Eye}, analogous to attention or working memory in humans, in which a set of items and their relationships are made explicit.  Examples of such relationships are occurrences of verbs that link a subject and an object, such as ``Odysseus meets Penelope'', and signifiers of context, such as, for example, that the story to follow takes place ``in the Bronze Age.'' 

This proposal describes a partial implementation of Robust Logic that fits well with and would easily augment current LLM technology. In this implementation the first step in both training and inference is to preprocess the text using a {\it semantic analyzer} \cite{Honnibal2017},\cite{Kamath2019},\cite{Wolf2019},\cite{SaparovMitchell2021} to detect relationships within sentences and a {\it discourse analyzer} \cite{Li2022} to detect these between sentences. The result will be (i) to augment the token set $V$ by a modest factor $g$ to $V'$, which will then be of size $|V'| = g|V|$, and (ii) to associate each token of text with $h-1$ further tokens from the augmented token set that give explicit information about the relationships between the objects to which that original token and other original tokens refer. In other words, an $N$ token sequence of text will be transformed into an $hN$ token sequence of augmented tokens. These added tokens facilitate the learning of Robust Logic rules and hence also reasoning. 

We call {\it Integracoded Machine Learning} such a practice of recoding the data to make it more complete for the purpose of enhancing certain behavior in machine learning. In the current instance the behavior being enhanced is reasoning while maintaining acceptable energy consumption. The terminology derives from the Latin word {\it integra}, meaning whole.

The argument therefore is that if the role of attention is narrowed to the essentials, then systems can be built that are more reliably correct in the respects of reasoning \cite{Wang2025}, factuality \cite{Wang2024} and absence of hallucinations \cite{Huang2024}. As far as factuality and absence of hallucinations, our expanded token set aims to capture relational information in the text more explicitly than does the text itself. To the extent that hallucinations and the lack of factuality in LLMs is caused by the LLM failing to internalize all the relationships that are implicit in the text, in either learning or inference, the more explicit recoding we suggest may help. It seeks to avoid the internalization of text at training or testing being merely impressionistic. As far as reasoning, correctly combining the implications of different parts of a text we call {\it chaining}. Robust Logic gives a principled basis for chaining information that has been learned and is therefore uncertain. Chaining in Robust Logic is provably {\it sound} in the sense that some probabilistic guarantees on the correctness of the conclusion follow from such guarantees on the constituents. For example, if each rule has been learned to 90\% accuracy, as statistically supported by samples from the training set, then some lower bound such as 80\% will be justified for the chained conclusion, assuming some stability on the probability distribution.

We will define a particular integracoding under which a core subset of Robust Logic rules becomes efficiently learnable in a defined sense.  Robust Logic offers a broad specification of {\it what we wish} to achieve. The implementation of the integracoded system described here can be regarded as a heuristic realization of it. While we are emphasizing linguistic inputs here, the methods may be applied more broadly, for example, in robotics, where the inputs would describe visual data and physical actions (e.g. \cite{Zeng2023},\cite{Zhong2025},\cite{Wang2026}).

The techniques used for LLMs, particularly transformers \cite{Vaswani2017}, have been the winners of a decades long competition between different heuristic learning algorithms in generalizing from large datasets. It is no surprise that what they accomplish well is generalization when there is a large data set available for statistical support. In the notion of educability \cite{Valiant2024}, which is aimed at capturing a broader set of human cognitive capabilities, such generalization is just one of three pillars. Another pillar is being able to learn by one-time instruction. For this it is important that the text of this instruction is well understood and does not need thousands of other similar instructions for disambiguation. Our enriched recoding must surely help in integrating knowledge acquired from such instruction with knowledge that has been learned from examples or acquired by instruction earlier. The third pillar of educability is the chaining of learned rules. Robust Logic is specially designed to make that principled. A more complete discussion of how chaining relates to broader notions of reasoning can be found in \cite{Valiant2024}.

\section{Robust Logic}

First, we describe by example how a window of text in an LLM is represented as a {\it scene} in Robust Logic. Suppose the text refers to three entities Bob, Joe and Sue, with the added information that Bob insulted Joe, that Sue likes Joe, and that Sue took revenge on Bob. This would be represented in a scene by the conjunction:
\begin{center}
Bob($x$) \& Joe($y$) \& Sue($z$) \& Insulted($x,y$) \& Likes($z,y$) \& Revenges($z,x$). 
\end{center}
The words Bob, Joe, Sue, Insulted, Likes and Revenges come from a fixed set of {\it attributes} that correspond to the tokens of LLM terminology.  The LLM will attempt to learn {\it rules} that hold widely for the scenes that are described in the training set of texts. 

An example of a {\it logical implication rule} is
\begin{center}
$\forall x \forall z [\exists y $ Insulted($x,y$) $\&$ Likes($z,y$) $\Rightarrow$ Revenges($z,x$) ]. 
\end{center}
\noindent
This means that if there is someone $y$ whom $x$ insulted and whom $z$ likes then $z$ will take revenge on $x$, whoever $x$ and $z$ are. However, the semantics of PAC learning is {\it not} such an implication. Instead it is an {\it approximate equivalence}, in which both positive and negative examples have to be correctly treated with high probability. A rule learned by Robust Logic will therefore be of the form of such an approximate equivalence:
\begin{center}
$\forall x \forall z$ $[\{\exists y$ DoesBadTo($x,y$) $\&$ Likes($z,y$)\} OR $\{...\}$ OR $\{...\}$  $\approxeq$ Revenges($z,x$)]. 
\end{center}

\noindent
Here the approximate equivalence sign $\approxeq$ occurs in place of the $\Rightarrow$ sign of standard logic. The important distinction is that now the left hand side has to satisfy the more onerous burden of being roughly equivalent to the right hand side, rather than being just one case of it, as is sufficient in an implication. This example suggests in two ways that greater coverage is needed for an approximate equivalence than for an implication. First, we used a higher level attribute ``DoesBadTo'' instead of ``Insulted''. Second, and, more fundamentally, we use a disjunction of several criteria on the pair of arguments $x,z$ of the right hand side ``Revenges'', rather than just one. (We note that Robust Logic requires only that the left-hand side in such an approximate equivalence rule belong to a learnable class. For this paper this ``OR'' form will be sufficient to consider.)

In Robust Logic the expressions allowed in each parenthesis pair ``\{...\}'' in the left-hand side of a rule are limited to a fixed set of {\it schemas}. An example of such a set is the following set of six schemas:\\
\begin{indent}
 (i)	$B(x)$, \\
 \end{indent}
 \begin{indent}
(ii)	$\exists y B(x, y)$, \\
 \end{indent}
 \begin{indent}
(iii)	$\exists y B(x, y) $\&$ C(y)$,\\
 \end{indent}
 \begin{indent}
(iv)	$\exists y B(x, y) $\&$ C(x, y)$,\\
 \end{indent}
 \begin{indent}
(v)	$\exists y \exists z B(x, y) $\&$ C(x, z)$,\\
\end{indent}
 \begin{indent}
(vi)	$\exists x B(x, y) $\&$ C(x, z)$.\\
\end{indent}
\noindent
The given example $\exists y$ DoesBadTo($x,y$) $\&$ Likes($z,y)$ is an instance of (vi) (after reordering and renaming of variables and attributes). It is implicit in this notation that the different symbols $x,y,z$, etc., refer to different objects in a scene, which translates to different positions in the text. {\it In our LLM implementation, we do not need to restrict the system to such a fixed schema set upfront!} Any rule will be learned that conforms to schemas that are simple enough to be learned with the available resources and are supported by sufficient numbers of example scenes in the training data. 

In the above example of a scene the relationships, such as Likes, refers to relationships between multiple entities in the world. Certain relationships such as {\it context relationships} may span large distances in the text. For example, a context indicator, {\it Bronze Age}, may have large span. Relationships can also be {\it linguistic} properties or relationships such as whether a token refers to a noun or not, or whether two tokens are close together in the text.  Several such relation types have been implemented in a Robust Logic based language model \cite{MichaelValiant2008}, but not with the integracode proposed here.

\section{The Unary Relational Integracode Makes Rules Learnable}

The class of Disjunctive Normal Form Boolean formulae, where each disjunct is the conjunction of at most $k$ literals is called $k$-DNF. The following is an example of 2-DNF:

\begin{center}
$x_1 \& x_3$ OR  $x_2 \& x_3$ OR $x_1 \& x_5$ OR $x_6$ OR $x_2 \& x_7$.
\end{center}

\noindent
It is known that this class of formulae is PAC learnable when $k$ is regarded as a constant \cite{Valiant1984}. (See \cite{KearnsVazirani1994} for general background.) We will show here that the task of learning a core subset of Robust Logic rules becomes an instance of learning $k$-DNF expressions after an appropriate recoding of the data. 

We now describe the {\it Unary Relational Integracode} or {\it URI} into which we propose to recode the text in order to make Robust Logic rules learnable during training and applicable to instances during inference. We propose a preprocessing stage of semantic and discourse analysis on the text and use the result of that to enlarge the token set and augment the text in both training and inference. The enlargement will consist of enlarging the token set $V$ of the basic text by a modest factor to $V'$, where the size of $V'$ is  $|V'| = g|V|$ and $V \subset V'$. The enlarged token set $V'$ will allow information about the relationships among the different token instances in the original text to be expressed more explicitly. The text augmentation then consists of adding $h-1$ {\it augmenting tokens}, now from the augmented token set $V'$, for each {\it original} token from $V$, to make a sequence of $h-1$ augmenting tokens in adjacent positions to each other and the original token.  In particular an $N$-token window of text will be expanded into a sequence of $hN$ tokens arranged in blocks of $h$ where in the $i^{th}$ out of the $N$ blocks, the first token $T_{i,1}$ would be the same original token (as used in a standard LLM), and $T_{i,2} ... T_{i,h}$ would be tokens from $V'$.

For example, consider a text, as above, that refers to a scene expressing the following:
\begin{center}
Bob($x$) \& Joe($y$) \& Sue($z$) \& Insulted($x,y$) \& Likes($z,y$) \& Revenges($z,x$).  
\end{center}

\noindent
Suppose that the names Bob, Joe and Sue occur in positions $i,j,m$, respectively, in the text, and correspond to $x,y,z$, respectively. Then $T_{i,1}$, $T_{j,1}$, $T_{m,1}$ would be the tokens Bob, Joe and Sue, respectively, from $V$. (If multiple tokens represent an entity some subset may be taken as the representatives, for example.) Each binary relation, such as Insulted, will be broken into two unary relations. For example, Insulted becomes Insulted$^1$ and Insulted$^2$, which respectively refer to the first and second argument of the binary relation Insulted, namely the subject and object. To express this we make $T_{i,2}$ = Insulted$^1$ and  $T_{j,2}$ = Insulted$^2$. In this way the semantic or discourse analyzer will tie the person doing the insulting to block $i$ and the person being insulted to block $j$. Similarly, perhaps, $T_{m,3}$ = Likes$^1$ and  $T_{j,3}$ = Likes$^2$. (As far as the ordering within the blocks, they may be placed in some fixed order in the order they are identified, or, alternatively, a hashing process may be tried.)

Now for each pair $p,q$, the token $T_{p,q}$ can be thought of as a sequence of $|V|$ Boolean bits if $q=1$ and $|V'|$ bits otherwise, with each bit representing the presence or absence of a particular original or augmenting token there. Then the totality of all the $T_{p,q}$ can be viewed as a bit sequence $S$ of length $N(|V|+(h-1)|V'|)$. 

The main observation is that, for the correspondence $x,y,z$ with positions $i,j,m$, respectively, the expression $\exists y $ Insulted($x,y$) $\&$ Likes($z,y$) is representable as a 4-DNF formula over this sequence of Boolean variables where each constituent 4-conjunction will be, for appropriate $i,j,m$,  of the $(i,j,m)$-form: 
\begin{center}
$[T_{i,2}$ = Insulted$^1$] $\&$ [$T_{j,2}$ = Insulted$^2$] $\&$ [$T_{m,3}$ = Likes$^1$] $\&$ $[T_{j,3}$ = Likes$^2]$.
\end{center}

\noindent
Note that we will be learning from inputs where Revenge$^1$ and Revenge$^2$ already occur as augmenting tokens in some blocks. We will be predicting Revenge$^1$ and Revenge$^2$ at points $i$ and $m$ for the example above, for inputs that are not yet so labeled. The Revenge$^1$ function that needs to be learned at $i$, therefore, is the disjunction over $j$ and $m$, of 4-conjunctions of the above $(i,j,m)$-form with that $i$ fixed, and also over the different locations in the $j$ and $m$ blocks where the Insulted$^2$, Likes$^1$ and  Likes$^2$ may be represented. In the same way, the Revenge$^2$ function that needs to be learned at $m$ is the disjunction over the $j$ and $i$ blocks and positions within them of 4-conjunctions of such $(i,j,m)$-forms with that $m$ fixed. These are 4-DNF formulae in both cases. (Note that the above notation is itself an abbreviation. For example,  $T_{i,2}$ really refers to a string of $|V'|$ Boolean variables, and ``$T_{i,2}$ = Insulted$^1$'' is the condition that the Boolean variable among these that corresponds to Insulted$^1$ will have value 1.) 

Now consider the following Robust Logic rule: 

\begin{center}
$\forall x \forall z$ $[\{\exists y$ Insulted($x,y$) $\&$ Likes($z,y$)\} OR $\{...\}$ OR $\{...\}$  $\approxeq$ Revenges($z,x$)]. 
\end{center}

\noindent 
We just showed that the expression in the first of its three $\{...\}$ subexpressions can indeed be represented by a 4-DNF expression in the Boolean variables. Now if the other two $\{...\}$ subexpressions in ``OR $\{...\}$ OR $\{...\}$'' are also from schema (vi) or from schemas that are no more complex, then these too will be so representable, and hence so will the contents of $[...]$, which we call the left hand side of the rule. Hence, Revenge$^1$ and Revenge$^2$ will also have 4-DNF expressions. (Note that we insist that the quantifications, such as $\exists y$, that occur in the various $\{...\}$ subexpressions are independent of each other.)

Note that it is the job of the semantic or discourse analyzer to determine if two mentions of Joe in the text, perhaps once by a pronoun, refer to the same person, and, if so, to ensure that in at least one of them, say block $j$, both attributes Insulted$^2$ and Likes$^2$ are present.

What can be claimed for the system described is that it will be able (i) to learn the needed rules given appropriate data and computational resources, and (ii) to predict correctly and at the appropriate places unary relations, such as Revenge$^1$ or Revenge$^2$, as implied by the rules. In particular, for learning, for each of Revenge$^1$ and Revenge$^2$, an appropriate DNF formula can be learned as supported by the dataset. If the Revenge relation occurs mostly for combinations of $(i,j,m)$ that are close in the text, then this will be reflected by having fewer terms in the learned 4-DNF, which will then be also sufficient at inference. For inference, suppose a text is input where the Revenge relation should be predicted at one or more pairs of places. This scheme handles all such cases correctly in the following sense: Predictions of Revenge$^1$ are made exactly at those values of $m$ such that there exist $i,j$ such that the 4-conjunction of the form $(i,j,m)$ is true, exactly as required. Note that here $i,j,m$ correspond to entities $x,y,z$ having the relations Likes and Insults as stated.  The analogous argument holds for Revenge$^2$. The scheme therefore correctly predicts the subjects and objects of the revenge even if there are multiple occurrences of such pairs, (but without disambiguating which is paired with which.)

One needs to be aware of the following complication. The particular integracode we have given hides some important information about the relations. If only one Likes relationship, such as ``Sue Likes Joe'' occurs in a certain vicinity of text then the connection between the occurrences of Likes$^1$ and Likes$^2$ is unique and above described DNF functions correctly. On the other hand, if the text contains multiple occurrences of the Likes relation with different subjects or objects, (i.e., repeated relations), then the DNF would be prone to false positive predictions, for example, of Revenge$^1$ for an additional person Joan who likes some further person Bill. However, the reason for this not being a concern is that what is being learned is not the DNF but some other function that has bee verified by testing to hold for real integracoded data and therefore not have the false positive problem. We are discussing the DNF only to show that the functions needed to describe the Robust Logic rules are PAC learnable. We do not and cannot suppose that these are exactly what the learning algorithm learns. What we are desribing is a heuristic partial realization of Robust Logic. (As an aside we note that the false positive problem that occurs with the DNF occurs only in text vicinities with repeated relations. These vicinities are easily detected and there, as an extreme measure, one has the option of not making relational predictions at all.)

We now observe that the composition of two rules from two possibly different schemas will also be expressible as $k$-DNF expressions, but with $k$ possibly larger than needed for either of the composed rules. For example, substituting the $D(y,z)$ from $\forall y \forall z \{\exists x B(x, y)C(x, z)  \approxeq D(y,z)\}$ into $\forall y \{\exists z D(y,z)F(z)  \approxeq E(y)\}$ gives  
\begin{center}
$\forall y \{\exists x \exists z B(x, y)C(x, z)F(z)  \approxeq E(y)\}$, 
\end{center}
which would require 5-DNF since conjunctions of $B^1,B^2,C^1,C^2$ and $F$ are needed.

We note again that no assumption about schema sets needs to be made upfront. The algorithm will learn rules that produce $k$-DNF formulae that are simple enough to be learned given the resources available to the system. In our earlier example of a schema set, all six schemas would give at most 4-DNF, while schema (ii) needs just a 2-DNF and schema (iii) a 3-DNF. 

In current LLMs there is an embedding that maps the $|V|$ dimensional space of the original tokens into a space of numerical vectors of some smaller dimension $d$. We expect that this can be adapted for our implementation also so that the $|V'|=g|V|$ dimensional space would map into $d'=dg'$ dimensions, say, for some $g'$. Hence, after accounting also for the size $h$ blocks we find that the length of the recoding will no longer be $dN$ but a factor of about $W=hg'$ larger. One would want $W$ to be a modest number.

To recap, how does one get sufficient expressive power with just unary relations? First, the set of relations that act on the same object in a scene in Robust Logic are linked in our Unary Relational Integracode by the proximity of being within one block. Second, the different unary relations, such as Likes$^1$ and Likes$^2$ that represent one original relation are linked by co-occurrence in the training set after the preprocessing. Note that any relation of any arity $r$ can be decomposed into $r$ unary relations, each acting on a single position, in exactly as the same way as the examples above for the relations Insults and Likes, which both have arity $r=2$.

Finally, let us state the general result that is established by the arguments of this Section. Consider expressions that conform to schemas, such as (i)-(vi) above, that are conjunctions of relations with existential quantification of some arguments. These we call {\it conjunctive expressions}. Each such conjunctive expression will correspond to $k$ unary relations of the integracode, where $k$ is the sum of the arities of all the occurrences of relations in the expression. We define a {\it Core Robust Logic} rule to be a rule where the left-hand side is a disjunction of such conjunctive expressions, where the existential quantifiers of each expression are independent of each other. We define the {\it maximum arity sum} of the rule to be the maximum of the arity sums of the constituent conjunctive expressions. Then the following has been established:

\begin{proposition} A Core Robust Logic rule of maximum arity sum $k$ has a corresponding $k$-DNF Boolean formula in terms of Unary Relational Integracode that correctly predicts the right-hand side of the rule on scenes that have no relation occurring more than once. 
\end{proposition}

As discussed above, we believe that the method developed here will be applicable to texts in general, not just those where each relation occurs just once in any vicinity.

Finally, to set the method in context we now compare it to the general results for Robust Logic \cite{Valiant2000}, which were implemented in \cite{MichaelValiant2008}. The main difference is that in the currently proposed integracode the expansion of the input text is by a factor of $h$. This we regard as a constant, imposed as a heuristic, and bounded above by the number of tokens that a single occurrence of a token may be usefully related to. Hence our integracode is a {\it succinct} representation of the input data in being of size linear in that of the data. In contrast, the general method used for encoding the relationships on a scene in \cite{Valiant2000} was a Booleanization of the relations that for relations of arity $r$ and a window of size $N$ created potentially $N^r$ new Boolean variables, and hence expanded the representation of the input data by such a factor. The operations are then polynomially bounded in terms of that exploded input representation, which, is quadratic already for binary relations and therefore does not scale practicably for large window sizes.

\section{Functional Advantages}
Most humans and, recently, most LLMs, can answer questions such as ``Did Aristotle own a coffee making machine?'' Neither needs to be trained on examples of Greek philosophers with their kitchen appliances. The capability, one presumes, for both humans and LLMs, is based on an ability to {\it chain} multiple rules that have been learned separately. Such chaining is a form of reasoning. Humans are able to perform this function reliably enough that one has to presume that there is an identifiable phenomenon at its core. Robust Logic aims to capture that phenomenon. It executes each step of chaining as an update on a scene in the Mind's Eye, using a rule that it has learned from the scenes it has encountered. The soundness of chaining follows as a mathematical consequence, under the assumption that the rules hold for the distribution of scenes even as the updating proceeds. 

The following are three advantages that an ability to learn Robust Logic rules can offer: 

\subsection{Functionality I: Applying a Learned Rule Within a Classifier}
If a Robust Logic rule can be learned during training, as described above, then at inference it can be applied. Thus a likely instance of revenge described in a text by having the words Likes and Insulted fitting the paradigm of our example rule, will be predicted by the learned classifier as an instance of revenge, with the unary relations Revenges$^1$ and Revenges$^2$ predicted at the appropriate places in the output.

\subsection{Functionality II: Chaining Multiple Learned Rules Between Classifier Calls}
One novel functionality that our architecture offers is that it can also {\it output} the augmenting tokens that give explicit information about the relationships in the text. This supports the chaining of rules {\it between} classifier calls. Currently, when successive tokens are predicted by an LLM, the successive calls to the classifier lose whatever relational information it may have gleaned internally in the previous call. Within a transformer the width of the feed-forward network is often wider than the embedded text length $dN$ by some factor, such as 4. It may be that the network seeks to discover internally similar information to what our Unary Relational Integracode expresses explicitly. Whether this is true or not, this information is certainly lost in the now standard architecture between different calls. Note that it has been shown in a small-scale experiment that such chaining can achieve superior missing word predictions \cite{MichaelValiant2008}. Our architecture uniquely incorporates this extra information between different calls to the learned LLM classifier. 

\subsection{Functionality III: Chaining Multiple Learned Rules Within a Classifier}
More speculatively, if multiple rules can be learned in the classifier at successive levels, so that, at inference, these rules can be applied in succession, then this would allow the possibility of principled chaining to be accomplished {\it even internally} within an end-to-end trained network. Of course, in that instance, the learned combinations might be viewed as basic rules themselves.

\section{Some Issues}
As in most applications of machine learning a big imponderable is the nature of the data source. For LLMs, for example, we have no characterization of what makes natural language texts as learnable, as they have been found to be, as far as the generation of well-formed sentences and paragraphs. Here we are seeking a functionality that is more challenging still and, again, we do not know the characteristics that would be most conducive to that. Certainly, there are data sources that we believe are not learnable, such as regular languages in worst case distributions\cite{KearnsValiant1994} or even on the uniform distribution \cite{Daniely2021}. Hence, any proposal, including this one, needs experimentation and empirical validation. Here we summarize the several questions that need exploring.

One issue is the nature of the $k$-DNFs formulae that need to be learned for natural language datasets. As a comparison, consider the problem of learning Boolean parities, namely the problem of predicting 0 or 1 for Boolean vectors where this value depends on whether the number of 1s in an unknown subset of the inputs is odd or even. There is ample empirical evidence that for this problem deep learning is good at reaching the limits of what theory allows. In the case of parities, Statistical Query learning algorithms, of which deep learning is an example, require training time exponential in $k$, the length of the parity being learned. It is known that conventional deep network training performs up to these limits \cite{Barak2022}.

The specific question here is whether the learning required is easier than general $k$-DNF learning, for which the best algorithms known for worst case formulae and worst case distributions is exponential in $k$. A factor that may help is that the different words that the relation Revenge links in the text often occur close together so that the number of $k$-conjuncts that need to be in the formula is much less than the worst case of $N^k$ where $N$ is the window size. It is more like $M^k$ where $M$ is the maximum distance in the text between any pair of linked items. When learning simple disjunctions attribute-efficient learning is possible where with resources that depend on the actual number of terms that occur in the particular disjunction rather than the worst case number \cite{Littlestone1988}. 

How much chaining of rules can be realized within a single end-to-end trained network call and how much needs to be done between different calls? We expect that some chaining will always be useful between calls, as discussed in \cite{Valiant2024}. The current practice of "thinking" or "chain-of-thought" reasoning are examples of chaining on language representations, between calls \cite{Wei2022},\cite{Zhang2022},\cite{Guo2025}. Our results show that the learning of Robust Logic rules that satisfy simple enough schemas should be feasible and hence that this kind of chaining will be supported, and will be more principled. What is less clear is whether such rules can be learned as a stack within the same network call, so that for example, the first half of the layers learn one rule, and the second half a second rule, which then compose at inference.  

Last but not least we need to determine the tradeoffs between the higher functionality we are seeking to achieve and the energy costs. In the context of transformers, the currently dominant architecture, if the window size of the the text presented to the learning algorithm is $N$ tokens and the dimension of the word embeddings is $d$, then the computational complexity and energy consumption of executing the standard transformer for one input at training or inference is O($dN^2$ +$d^2N)$, where the first term comes from the self-attention layers and the second term from the feed-forward layers. There have been many proposals to reduce one or the other term in this quantity \cite{Beltagy2020},\cite{Choromanski2020},\cite{Kitaev2020},\cite{Wang2020},\cite{Hoefler2021},\cite{Gu2021},\cite{Hoffman2022},\cite{Tay2022},\cite{Sun2025}, such as by using sparse rather than dense matrices. 

Now let us suppose that, in accordance with our theory, the difficulty of training is substantially reduced by our explicit recoding. Then there may be a substantial energy saving if learning algorithms of lower complexity, such as sparser forms of transformers, suffice. Ideally one aims for computational complexity of O($dN$), without the extra factors of $d$ or $N$.

As far as the overheads of our method, if the text window processed at a time by the learning algorithm is $N$ original tokens, then our architecture will widen the number of tokens to be processed to $hN$ where $h$ is the block size. Further, since we are augmenting the token set from $V$ to $V'$, by a factor of $g$, the embedding dimension of the tokens can be expected to increase also, from $d$ to $g'd$, say. Hence, if the linear O($dN$) complexity is achieved, because of the more explicit input recoding, then our architecture would have complexity O($g'hdN$) instead of the nominal O($dN^2$ +$d^2N)$. This would be a saving in the case that $g'h \ll d,N$, which is what we are aiming for. In widely used systems, $d$ is at least in the thousands and $N$ even larger, while for $h,g'$ we are aiming for upper bounds of, say, 10 or perhaps rather less. (Note that one might also consider schemes where a whole token block is embedded directly into a space of dimension $g^*d$, in which case $g^*$ would replace $hg'$ in the analysis above.)

On the negative side of the energy ledger, we have semantic and discourse analyzers to perform some preprocessing on the input text. One expects that useful such analysis can be done at an energy cost that is no more than a small fraction of the energy costs of a learning network in training or inference. This question is, of course, intimately related to that of determining what analysis should be done. An earlier laptop-scale investigation of natural language prediction with Robust Logic (but not the currently proposed Unary Relational Integracode) found that some computationally efficient analyzers did improve the accuracy of predicting missing words in text, a task very close to the central task of current LLMs \cite{MichaelValiant2008}. One example of relations used there was subject-verb-object, much as in the ``Bob insulted Joe'' example above. This we expect to be very useful in general. A second example of relations used in that study is one of proximity, indicating whether a certain word is within a certain distance of another in the text. A third example was tagging words with their part of speech. A further example not pursued there, but no doubt very useful, is establishing coreferences among different sentences and then detecting subject-verb-object relations. The question of which of these and alternative possibilities that semantic and discourse analysis can provide are most useful as augmenting tokens in our architecture remains a task for investigation. A related question is the accuracy of the analysis needed. 

\section{Conclusion}
It may well be that computationally feasible generalization from examples is the most spectacular cognitive phenomenon that will ever be identified and will remain the foundational basis of AI. However, it does not follow that further ideas will not be useful. 

Robust Logic can be regarded as a method for realizing a symbolic world model that facilitates reasoning that is supported by provable soundness arguments. The soundness arguments apply to the chaining of rules that are learned from data. {\it These are soundness arguments for which no known analogs apply to current LLM practice, where the chaining is performed directly on segments of text.} Here we have proposed a transformation on input data that offers a heuristic for realizing a core subset of Robust Logic.

The Unary Relational Integracode proposed here simultaneously satisfies three fortuitous and perhaps surprising quantitative properties. First, it is succinct in terms of the encoded data. Second, the relational learning task demanded of it, namely the learning of the rules of Core Robust Logic, is from a provably learnable class in the sense described. Third, it enables a single classifier with many outputs to learn rules for them simultaneously. This illustrates the principle of Integracoded Machine Learning, where a recoding that makes the data more whole is used to facilitate a particular behavior, in this case better reasoning with limited energy overheads. Previously, data recoding in general may have been thought to have limited promise because of the belief that generic machine learning algorithms directly applied to raw data are hard to beat for generalization. Where effective data is essentially unlimited or can be artificially generated this belief has some merit, but here we are seeking to enhance reasoning, which is not entirely a large-data phenomenon. Reasoning is most vital for novel situations where experience of similar ones is lacking.

\bibliographystyle{plain}
\bibliography{Valiant2026Xbib}
\end{document}